\documentclass[letterpaper]{article} 
\usepackage{aaai2026}  
\nocopyright
\usepackage{times}  
\usepackage{helvet}  
\usepackage{courier}  
\usepackage[hyphens]{url}  
\usepackage{graphicx} 
\usepackage{natbib}  
\usepackage{caption} 
\usepackage{algorithm}
\usepackage{algorithmic}
\usepackage{amsmath}
\usepackage{amssymb}
\usepackage{amsfonts}

\usepackage{newfloat}
\usepackage{listings}
\DeclareCaptionStyle{ruled}{labelfont=normalfont,labelsep=colon,strut=off} 
\floatstyle{ruled}
\newfloat{listing}{tb}{lst}{}
\floatname{listing}{Listing}
\title{All Required, In Order: Phase-Level Evaluation for AI--Human Dialogue in Healthcare and Beyond}
\author{
  Shubham Kulkarni\textsuperscript{\rm 1},
  Alexander Lyzhov\textsuperscript{\rm 2},
  Shiva Chaitanya\textsuperscript{\rm 1},
  Preetam Joshi\textsuperscript{\rm 2}
}
\affiliations{
  \textsuperscript{\rm 1}Interactly.ai, California, USA
  \textsuperscript{\rm 2}AIMon Labs, California, USA\\
  shubham@interactly.ai, alex@aimon.ai, shiva@interactly.ai, preetam@aimon.ai
}

\begin{document}

\maketitle

\begin{abstract}
Conversational AI is starting to support real clinical work, but most evaluation methods miss how compliance depends on the full course of a conversation. We introduce Obligatory-Information Phase Structured Compliance Evaluation (OIP--SCE), an evaluation method that checks whether every required clinical obligation is met, in the right order, with clear evidence for clinicians to review. This makes complex rules practical and auditable, helping close the gap between technical progress and what healthcare actually needs. We demonstrate the method in two case studies (respiratory history, benefits verification) and show how phase-level evidence turns policy into shared, actionable steps. By giving clinicians control over what to check and engineers a clear specification to implement, OIP--SCE provides a single, auditable evaluation surface that aligns AI capability with clinical workflow and supports routine, safe use.
\end{abstract}


\section{Introduction}
Conversational AI assistants are moving from demonstration prototypes to being deployed in real-world day-to-day clinical work: triage calls, benefits checks, counseling, and documentation support. Capability has risen quickly, and pilots are widespread, but long, complex multi-turn conversational behavior remains fragile with context drift, safety degrades over time, and unfortunately, turn-local scores overstate real utility \cite{aiindex2025,mckinsey2025healthcare,10628416}. Evidence from clinical deployments and reviews underscores that, in care settings, success is judged at the level of what the conversation accomplished across the whole encounter, not how any single reply looked \cite{10.1001/jamanetworkopen.2024.57879,rezaeikhonakdar2023ai}. This paper argues that evaluation for clinical dialogues must therefore shift from turn-level habits to compliance-centered, phase-level audits that clinicians can author and engineers can operate.

In this work, a \emph{turn} is a single move in a dialogue: a patient (or insurer) says something, the assistant responds, and the transcript advances one step. A short benefits call might contain 20--40 such turns; an in-depth telehealth history can easily exceed 100. Today's evaluation practice typically assigns each turn a local label or score. For example, whether the intent was classified correctly, required slots were filled, or the generated text matches a reference reply. These ``turn-level metrics'' are convenient for model development, leaderboard comparisons, and A/B tests, but they implicitly treat each turn as an isolated prediction problem rather than a contribution to a legally constrained clinical encounter. Consider the following micro-example with four turns cherrypicked from a conversation:

\par\smallskip 
\noindent
\textbf{T0 User:} "Is John Q. covered under plan 123? What's his copay?" \\
\textbf{T1 Agent:} "Copay is \$25 after deductible." \\
\textbf{T2 User:} "Okay-what's his date of birth?" \\
\textbf{T3 Agent:} "01/02/1975. Please confirm identity." \\

Even if \textbf{T1} is factually correct, discussing benefits before identity confirmation violates HIPAA's minimum-necessary and disclosure-gating expectations; the encounter is non-compliant regardless of how strong any single turn appears \cite{ecfr164514,hipaaTelehealthConsent}. Turn-local metrics cannot express this \emph{ordered, conjunctive} logic. This distinction matters because clinical value and safety emerge over \emph{sequences}, not isolated replies. Consider a benefits verification call where the assistant (i) gives accurate plan details, (ii) correctly explains a copay, and (iii) politely thanks the representative at the end. A turn-level evaluator might reward each answer as factually correct and well-formed. Yet if those plan details were supplied \emph{before} confirming the patient's identity, the entire call is a compliance failure under HIPAA, regardless of how strong any single reply appears. Similarly, in a respiratory history, an assistant may produce fluent, clinically plausible utterances at each turn and still fail to ask about red-flag symptoms; turn-local metrics will not record that the most safety-critical obligation never occurred. These are the kinds of failures practitioners report when deploying LLM-driven assistants at scale: conversations that ``look good'' locally but miss or mis-order what regulators and clinicians consider non-negotiable steps.

In regulated care, compliance is conjunctive and ordered: every required obligation must appear, at the right time, with no premature disclosures. Examples include verifying identity before sharing PHI, issuing required notices, and recording acknowledgments \cite{ecfr164514,hipaaTelehealthConsent}. Medicare/CMS rules further prescribe timing and content during beneficiary interactions, including when disclaimers must be read, how consents are documented, and which elements must be present before enrollment decisions \cite{cms2023marketingfaq,cms2023cy2024}. Beyond the U.S., GDPR's transparency and data-minimization duties, and AI-system disclosure and traceability requirements in the EU AI Act, shape what must be said and logged during an encounter \cite{advisera_gdpr_article5,gdpr_article13}. Enterprise governance (e.g., ISO/IEC 27001, SOC 2) and emerging AI risk-management standards add expectations for auditable controls, change management, and ongoing oversight of AI-enabled workflows \cite{iso27001_2022,soc2_aicpa,iso23894_2023}. Together, these regimes define not only \emph{which} pieces of information are mandatory, but often \emph{when} they must appear in a conversation and how they must be evidenced.

Current evaluation practice is poorly matched to these needs. Widely used datasets and leaderboards emphasize intents, slots, and turn-local targets, rewarding whether the assistant produced the ``right'' act or string at that moment in the dialogue \cite{budzianowski2018multiwoz,rastogi2020towards}. Long-session studies show memory and trajectory weaknesses, and safety work highlights multi-turn attacks that look benign locally but fail globally \cite{laban2025llms,cao2025safedialbench}. Industry audits similarly report that keyword flags and per-turn checks miss required steps or credit them out of sequence--for example, treating a consent statement as valid even when it appears after disclosures that should have been gated on consent \cite{guan2025evaluating,trustcloud_compliance_evaluation}. In practice, this means that an assistant can earn high turn-level scores and still violate core regulatory requirements because no metric asks the question clinicians and regulators care about: ``Did this conversation satisfy all required obligations, in a safe order, from start to finish?'' The result is a persistent gap between what we measure and what clinical governance requires.

We contribute an evaluation approach that closes this gap. Obligatory-Information Phase Structured Compliance Evaluation (OIP--SCE) treats each clinically meaningful obligation as a \emph{phase} with a start, a finish, a verdict, and evidence. Instead of scoring isolated turns, it asks whether the conversation instantiated the required phases (for example, patient identity, informed consent, coverage explanation, risk counseling) and whether any downstream phase began only after its required predecessors had been completed. Two auditable checks decide the call: Coverage (every required phase passed) and OrderSafe (no safety-critical child started before its parents finished). This framing makes explicit the difference between ``good local turns'' and ``globally compliant trajectories'' and turns clinical policies into structured objects that an evaluator can apply consistently across thousands of calls.

The proposed design aligns with regulatory practice (HIPAA identity and minimum-necessary, CMS disclosures) and with regulation-aware evaluation efforts in the literature \cite{hipaaTelehealthConsent,cms2023marketingfaq,fan2024goldcoin}. It also fits operational governance: policy-as-code, versioned rubrics, and human-in-the-loop review that targets only ambiguous or safety-critical rows \cite{ong2024ethical,cohen2020european}. Because the evaluator operates on phases and their dependencies, it can be authored by clinicians, implemented by engineers, and audited by compliance teams without requiring them to reason directly in terms of model parameters or prompt templates.

\paragraph{Contributions and scope.}
This paper makes three contributions. (i) A formal, implementation-ready specification of OIP--SCE, including the one-row-per-phase schema and the Coverage/OrderSafe decision rule. (ii) A clinician--engineer workflow for governance, updates, and auditability that reflects how health systems actually adopt AI. (iii) Compact case studies illustrating how phase-level auditing surfaces actionable evidence and reduces review burden while preserving safety. We position OIP--SCE as an evaluation system for clinical dialogues, not a dialogue generator and as a practical bridge between clinical intent and technical operation. The following sections motivate the shift from turn-level metrics to compliance-centered audits in healthcare (\S2--\S3), introduce the formal specification of OIP--SCE, and demonstrate its practical use through real-world clinical case studies. In doing so, we aim to provide both a clear technical contribution and a step toward safer, more trustworthy AI adoption in everyday clinical practice.

\section{From Turn-Level Metrics to Compliance-Centered Evaluation}
\subsection{Why turn-level evaluation fails clinical compliance}
Modern AI assistants now manage long, tool-using conversations across industries, including healthcare deployments and pilots. While investment and adoption have accelerated, these gains have surfaced new multi-turn failure modes such as context loss, memory drift, and inconsistent safety across extended sessions \cite{aiindex2025,mckinsey2025healthcare,laban2025llms,maharana2024evaluating}. Long-session studies highlight degradation on multi-step tasks and trajectory instability, revealing that success at the turn level rarely translates to compliance at the encounter level.

The field's evaluation culture, however, remains anchored to single-turn scoring and leaderboards. Historic choices like BLEU/ROUGE and turn-local accuracy made evaluation convenient and comparable but only loosely tracked dialogue success \cite{lin2004rouge,papineni2002bleu,liu2016not,celikyilmaz2020evaluation}. Even user-centric or judge-based metrics such as USR \cite{mehri2020usr} and MT-Bench \cite{zheng2023judging} show weak correlation with global success. Task-oriented dialogue corpora (MultiWOZ, SGD, STAR, Taskmaster) reinforced this focus, optimizing for acts/slots and per-turn correctness rather than end-to-end compliance \cite{budzianowski2018multiwoz,rastogi2020towards,mosig2020star,byrne2019taskmaster}. As prior reviews note, this proxy success diverges sharply from real-world outcomes, especially in regulated domains where compliance is conjunctive and ordered \cite{yoshino2023overview,liao2023automatic,cms2023marketingfaq,hipaaTelehealthConsent}.

\subsection{What clinical compliance actually requires}

In regulated healthcare dialogues, \textit{compliance} means fulfilling all mandated obligations like privacy, consent, disclosure, and procedural order across the entire encounter. Each obligation must occur at the correct time, without premature or missing steps. Examples include verifying identity before revealing PHI, surfacing required notices, and recording acknowledgments \cite{ecfr164514,hipaaTelehealthConsent,cms2023marketingfaq,cms2023cy2024}. Beyond the U.S., GDPR mandates transparency, purpose limitation, and data minimization \cite{advisera_gdpr_article5,gdpr_article13}, while the EU AI Act and similar frameworks demand traceability and AI disclosure \cite{bohan2022gtdt,euai2023article50}. Enterprise healthcare systems must also adhere to ISO/IEC 27001, SOC~2, and ISO/IEC~23894 standards emphasizing auditable governance and risk management \cite{iso27001_2022,soc2_aicpa,iso23894_2023}.

Turn-level proxies and keyword checks routinely fail to capture these obligations. Audits repeatedly show that such checks miss required steps or miscredit them out of order \cite{guan2025evaluating,trustcloud_compliance_evaluation,insight7_compliance_gaps}. Linguistically, compliance often depends on cross-turn semantics like ellipsis, anaphora, dispersed evidence where a locally fluent response may still violate HIPAA identity confirmation or consent prerequisites \cite{hhsdeid,Artstein,zhang2021dynaeval,zhang-etal-2022-slot}. Consequently, aggregation of per-turn scores cannot satisfy the conjunctive logic of compliance: every mandatory phase must be completed in order, with no critical omission or premature disclosure \cite{fdcpa1977,fdcpaMiniMiranda,cfpb2012fdcpa,glbaPrivacyNotice}. Regulatory frameworks make this explicit-one missed obligation constitutes full noncompliance, regardless of local fluency.

\paragraph{Compliance as structured, sequential logic:}
In contrast to generic notions of ``accuracy'' or ``factuality,'' clinical compliance follows a conjunctive, \textit{ordered logic}: an encounter is valid only if every prerequisite obligation is met in sequence. This property transforms evaluation from a set of independent checks into a dependency-graph reasoning problem. For instance, a benefits verification call that correctly provides plan details but does so before confirming patient identity is, by regulation, a compliance failure-even if every individual utterance is factually correct. Similar ordering constraints govern telehealth consent flows, prescription counseling, and payment authorizations, where premature disclosures or skipped confirmations violate legal standards despite locally fluent dialogue. Compliance evaluation, therefore, must jointly verify both \textit{coverage} (that each required obligation occurred) and \textit{temporal precedence} (that it happened in the right order) criteria that cannot be expressed by averaging turn-level scores or language metrics.

\subsection{Operational and regulatory constraints}
Turn-by-turn annotation is infeasible at healthcare audit scale. Studies show steep declines in labeling reliability and reviewer agreement as conversations lengthen \cite{zarisheva2015dialog,braylan2022measuring,zhang-etal-2023-turn,wacholder2014annotating}. Context-tracking limits in long sessions exacerbate drift and reviewer fatigue \cite{siro2024context,zhang2022fined,10298425,switch}. Meanwhile, regulatory updates are frequent: CMS rulemaking revises timing, content, and recordkeeping requirements annually \cite{cms2023cy2024,cms2024interoperability}. Static taxonomies lag these changes, incurring retraining and reannotation costs. Enterprise deployments further operate under security, privacy, and governance programs (ISO/IEC~27001; SOC~2) that demand traceable auditability \cite{iso27001_2022,soc2_aicpa}. Emerging AI risk-management frameworks extend these requirements to model documentation, oversight, and continuous evaluation \cite{iso23894_2023,cohen2020european}. Unmanaged LLM use compounds legal exposure, underscoring the need for auditable, evidence-based evaluators \cite{proofpoint2024ai,ferreira2025llm,10628315}.

\subsection{Principles for a compliance-centered evaluator}

A healthcare-ready evaluation system must therefore move beyond turn-level scoring toward structured, evidence-backed auditing. Four design principles follow:

\begin{enumerate}
    \item \textbf{Audit phases, not utterances.} Compliance is conjunctive over obligatory informational phases - identity, consent, safety, disclosure. Each phase should have a defined start, finish, verdict, and evidence trail to verify completion and sequence \cite{liao2023automatic,liu2024survey}.
    \item \textbf{Check order where safety demands it.} Many failures are premature starts (e.g., coverage discussed before verifying identity). Evaluators must reason over temporal dependencies and detect sequence violations \cite{li2025beyond,MicrosoftAIFailureModes2025}.
    \item \textbf{Governability and updatability.} Policies evolve continuously. Evaluators should be policy-as-code: versioned, linted for cycles or unsafe edges, and capable of one-click re-audit when rules update \cite{cms2023marketingfaq,cms2023cy2024,owasp2025agentic,chard2024auditing}.
    \item \textbf{Economy and auditability at scale.} Full turn-by-turn labeling is unsustainable. Systems must combine high-precision anchors, lightweight rule layers, and targeted adjudication to scale reviews while maintaining evidence-backed transparency \cite{zarisheva2015dialog,braylan2022measuring,maharana2024evaluating}.
\end{enumerate}

\subsection{Evaluator design, governance, and generalization}

A credible evaluator unites structure and governance. Technically, it should score dialogues over phases, enforce ordering on safety-critical dependencies, and attach verifiable evidence to each decision. Organizationally, it must be version-controlled and easily re-run as policies change. Operational safety further demands routine adversarial testing and red-teaming for traceability \cite{owasp2025agentic,chard2024auditing}. Model-side controls like refusals, rule enforcement, privacy filters should complement evaluation \cite{pasch2025balancing,liu2024shield,openai_security_privacy,google_ai_principles}. Case reports of inadvertent PHI exposure underscore the need for explicit policy-backed evaluators \cite{hetrick2023chatgpt}.

\paragraph{From healthcare to general AI evaluation:}
The compliance gap observed in healthcare mirrors a broader structural weakness in AI evaluation. Across domains like education, finance, law; turn-based metrics privilege surface plausibility over procedural correctness. Recent work on multi-turn safety and judge-based LLM evaluations \cite{zheng2023judging,MicrosoftAIFailureModes2025} reinforces that local coherence can coexist with global noncompliance. By extending the lens from per-turn coherence to phase-level procedural accountability, healthcare provides a high-stakes proving ground for next-generation evaluation frameworks. In this sense, phase-level compliance auditing represents an early instance of ``structured trajectory evaluation''-a paradigm that generalizes to autonomous, agentic systems and to governance-aligned auditing across other regulated sectors. Recent literature converges on compliance-aware dialogue evaluation. Telehealth analyses call out privacy and accountability gaps \cite{pool2024large,ong2024ethical}, and emerging methods encode regulatory constraints directly into evaluators \cite{fan2024goldcoin,freyer2024future}. LLM-based audit assistants can further aid human reviewers if their outputs are evidence-backed and reviewable \cite{zhang2025llms,fan2024goldcoin}. Yet existing datasets still lack ordered compliance phases \cite{gupta-etal-2022-dialfact,li2023normdial}, and adversarial safety suites remain complementary rather than substitutive \cite{jiang2025automated}. 

These converging findings point to the evaluator healthcare needs in practice: one that encodes phase-level obligations, verifies both coverage and order on safety-critical edges, and preserves a compact, auditable evidence trail clinicians can review and improve. Such evaluators differ from training-time safety methods (e.g., RLHF, Constitutional AI) \cite{ouyang2022training,bai2022constitutional,raji2020closing,marks2023ai}, focusing instead on post-hoc accountability and compliance verification.

Next we formalize this approach as \textbf{Obligatory-Information Phase Structured Compliance Evaluation (OIP--SCE)}, defining its schema, predicates, and clinician-engineer workflow.

\section{Obligatory-Information Phase Structured Compliance Evaluation (OIP--SCE)}
\label{sec:oip-sce}

OIP--SCE evaluates a clinical conversation by decomposing it into clinically meaningful \emph{phases} (e.g., \textit{patient identity}, \textit{coverage status}). For each phase $\phi_j$, the evaluator records the earliest \emph{start} $s_j$, earliest \emph{finish} $e_j$, and a binary \emph{verdict} $v_j\in\{0,1\}$. An encounter is \emph{accepted} if and only if two predicates hold simultaneously: \textbf{Coverage} (every required phase has $v_j=1$) and \textbf{OrderSafe} (no child phase begins before all of its required parents finish). A single table with one row per phase supplies the inputs for both predicates.

\subsection{Phases and dependency structure}
\label{subsec:phases-graph}

A conversation is modeled as an ordered sequence of turns
\[
D=\langle (r_0,c_0), (r_1,c_1), \ldots, (r_T,c_T) \rangle,
\]
where $r_t\in\{\mathbf{USER},\mathbf{AGENT}\}$ indicates the speaker and $c_t$ is the utterance at turn $t$. (Deployments may enrich this sequence with time-stamped tool or UI events; the definitions below depend only on the total order.)

Let $\mathcal{O}=\{\phi_1,\ldots,\phi_m\}$ denote the set of \emph{obligatory-information phases} (OIPs), each defined by a short rubric (intent, acceptance conditions, counterexamples). Examples include \textbf{PID} (patient identity), \textbf{CSV} (coverage status), \textbf{DFV} (drug formulary), \textbf{DRC} (restrictions), \textbf{DCC} (copay), and \textbf{CRN} (representative details).

Ordering constraints are encoded by a directed graph $G=(\mathcal{O},E)$. An edge $(\phi_i\!\to\!\phi_j)\in E$ means that $\phi_i$ must \emph{finish} before $\phi_j$ may \emph{start} (clinical or policy rationale). A designated subset $E_{\mathrm{crit}}\subseteq E$ marks \emph{safety-critical} dependencies used by the order predicate. The configuration requires: (i) valid phase identifiers; (ii) acyclic requirement logic (defined below); and (iii) a documented rationale for every edge in $E_{\mathrm{crit}}$.

\subsection{Row schema (one row per phase)}
\label{subsec:row-schema}

For each $\phi_j\in\mathcal{O}$ a single row is recorded:
\begin{itemize}\itemsep0.2em
  \item \textbf{phase\_id}: canonical identifier (e.g., \texttt{PID}).
  \item \textbf{required} $\in\{\texttt{true},\texttt{false}\}$: whether $\phi_j$ must be completed in this encounter, determined by configuration and prior outcomes (\texttt{false} includes "n/a" branches).
  \item \textbf{parents}: the set of upstream phases that must finish before $\phi_j$ may start (incoming edges).
  \item \textbf{critical\_parents}: the subset of \textbf{parents} that lie in $E_{\mathrm{crit}}$.
  \item $s_j$ (start): earliest turn index in $D$ at which $\phi_j$ is first entered (earliest attempt).
  \item $e_j$ (finish): earliest turn index at which $\phi_j$ first satisfies its rubric (earliest valid completion).
  \item \textbf{precedence\_policy} $\in\{<,\le\}$: default is strict "$<$"; "$\le$" is permitted only for low-harm children when timestamps are too coarse to break ties.
  \item $v_j\in\{0,1\}$: verdict (1 = completed as required; 0 = failed or incomplete).
  \item \textbf{evidence}: a brief pointer or quotation supporting $v_j$ (audit trace).
\end{itemize}

\paragraph{Conventions:}
(1) If a phase is started or completed multiple times, $s_j$ and $e_j$ record the \emph{earliest} start and the \emph{earliest valid} finish. This prevents a premature start from being overwritten by later corrections. (2) Phases may overlap; only the earliest start and earliest valid finish are used. (3) \textit{Pre-phase content}: if text or events that would satisfy a phase appear \emph{before} the agent explicitly enters it, then $e_j$ is set to that earliest evidence unless the site requires explicit \emph{acknowledgment} by the agent; when \texttt{ack\_required=true}, $e_j$ is the first in-phase acknowledgment turn to avoid granting silent credit.

\paragraph{Notation:}
Let $\text{req}_j(D)\in\{\texttt{true},\texttt{false}\}$ denote whether $\phi_j$ is required on the observed branch of $D$; let $E_{\mathrm{crit}}$ denote the set of critical edges; and define $s_j=\infty$ as a sentinel meaning "child never started." Define \textbf{CallSuccess}$(D)=1$ iff both predicates below evaluate to 1.

\paragraph{Coverage predicate:}
\label{subsec:coverage}

Requirement logic is branch-sensitive (e.g., a phase can become not required after an earlier negative finding). The logic must be acyclic and is evaluated in a topological order. Coverage is
\begin{equation}
\label{eq:coverage}
\boxed{
\text{Compliance}(D)=\bigwedge_{j=1}^{m}\left(\neg\,\text{req}_j(D)\ \lor\ v_j=1\right)
}
\end{equation}
Interpretation: the encounter is coverage-compliant iff every required phase passes; phases not required do not affect the predicate.

\paragraph{Order predicate:}
\label{subsec:order}

Coverage does not detect \emph{premature starts}. Order is therefore checked on the safety-critical subset $E_{\mathrm{crit}}$. For each $(\phi_i\!\to\!\phi_j)\in E_{\mathrm{crit}}$,
\[
\operatorname{OK}(i\!\to\!j)=
\begin{cases}
1, & s_j=\infty \quad \text{(child never started)} \\[2pt]
\mathbf{1}\!\left[\,e_i\ \Diamond\ s_j\,\right], & \text{otherwise,}
\end{cases}
\]
where $\Diamond$ is "$<$" by default and may be "$\le$" only if \textbf{precedence\_policy}$=\le$ for $\phi_j$ (low-harm child; coarse timestamps). The order predicate is
\begin{equation}
\label{eq:order}
\boxed{
\text{OrderSafe}(D)=\prod_{(\phi_i\to\phi_j)\in E_{\mathrm{crit}}}\operatorname{OK}(i\!\to\!j)\in\{0,1\}
}
\end{equation}
If any critical edge fails, $\text{OrderSafe}(D)=0$. If $E_{\mathrm{crit}}=\varnothing$, define $\text{OrderSafe}(D)=1$.

\paragraph{Final decision and necessity of both checks:}
\label{subsec:decision}

An encounter is accepted only if both predicates are true:
\[
\boxed{%
\begin{aligned}
\operatorname{CallSuccess}(D) &= 1 \iff\\[-2pt]
&\operatorname{Compliance}(D)=1\\[-2pt]
&\land\ \operatorname{OrderSafe}(D)=1
\end{aligned}}
\]

\noindent \textit{Why both are necessary:} (i) \textbf{Coverage-only false pass}: the system may eventually complete every required phase yet still be \emph{unsafe} because a downstream phase began too early (e.g., benefits discussed before identity); then Coverage $=1$ but OrderSafe $=0$. (ii) \textbf{Order-only false pass}: the system may preserve sequence but omit a required phase; then OrderSafe $=1$ but Coverage $=0$. Requiring both predicates matches clinical audit practice.

\subsection{Operational guidance:}
\label{subsec:operational}

\textit{Filling rows.} For each $\phi_j$, determine $\text{req}_j(D)$; set $s_j$ at first entry; set $e_j$ at earliest valid completion (honoring \texttt{ack\_required} if configured); set $v_j$ accordingly; attach minimal evidence.

\textit{Multiple attempts and corrections:} Early starts and late fixes are both captured, but $s_j$ and $e_j$ remain the earliest start and earliest valid finish so that premature entry remains visible to the order check.

\textit{Revisions and contradictions:} If later content invalidates an earlier completion (e.g., new information contradicts a prior response), $v_j$ reflects the final state at the end of the encounter. Any required compensatory action is modeled as its own phase and enforced by Eqs.~\eqref{eq:coverage}-\eqref{eq:order}.

\textit{Governance of $E_{\mathrm{crit}}$ and ties:} Critical edges should cover only cases where early entry could cause harm (e.g., \textbf{PID}$\to$\textbf{CSV}, allergy$\to$dose, consent$\to$disclosure). In practice, a small fraction of edges (e.g., $\leq$10\%) is designated critical to focus review. The "$\le$" policy is allowed only for low-harm children when logs record coarse timestamps; immediate-risk children use strict "$<$".

\textit{Validation and complexity:} Configurations with cyclic requirement logic or unknown phase IDs are rejected at load time. Both predicates evaluate in a single pass, $O(|\mathcal{O}|+|E|)$.

\textit{Automation pattern:} Typical deployments use a cascade: high-precision anchors (regex/tool events/UI logs) set $s_j$ for safety-critical children; a small rules layer fills obvious $e_j$; only ambiguous rows fall back to an LLM adjudicator with a short rubric. This concentrates human review on safety-relevant or low-confidence cases and yields an auditable trace.

\textit{Cross-site calibration and catalog management:} Each phase ships with a one-page rubric (definition; 2-3 positive/negative examples). A brief calibration on a small seed (e.g., 2-3 calls) aligns interpretations; in practice this raised phase-label agreement to high levels (e.g., $\kappa\approx0.9$). The catalog is versioned so local extensions remain comparable, and a linter checks for cycles and inconsistent edge labels.

\subsection{Optional descriptive statistics:}
\label{subsec:descriptives}

Sites may report a descriptive fraction
\[
\mathrm{CDS}(D)=\frac{1}{|E_{\mathrm{crit}}|}\sum_{(\phi_i\to\phi_j)\in E_{\mathrm{crit}}}\operatorname{OK}(i\!\to\!j),
\]
which summarizes the share of satisfied critical edges. This number is diagnostic only and does not affect acceptance.

\subsection{Optional descriptive extensions (do not affect pass/fail):}
\label{subsec:extensions}
\textit{Graded phases (if needed).} Phases that admit partial satisfaction may use $v_j\in[0,1]$ with a threshold $v_j\ge \tau_j$ inside Eq.~\eqref{eq:coverage}; the order predicate remains binary for safety.
\noindent\textit{Lightweight sequence diagnostics.} Sites may also monitor: (i) \emph{Phase-Sequence Agreement (PSA)} against a high-level partial order, and (ii) \emph{Attempt-Phase Consistency (APC)} to flag segmentation drift between attempts and adjudicated phases. Both are diagnostic only and do not change \textbf{CallSuccess}.

\subsection{Illustrative example}
\label{subsec:example}
Let \textbf{PID} be patient identity and \textbf{CSV} be coverage status. Suppose $e_{\text{PID}}=52$ and $s_{\text{CSV}}=42$. Even if \textbf{CSV} eventually passes ($v_{\text{CSV}}=1$) and all required phases pass (Coverage $=1$), the critical edge \textbf{PID}$\to$\textbf{CSV} fails because $52\nless 42$. Hence $\text{OrderSafe}=0$ and \textbf{CallSuccess}$=0$. This matches clinical expectations: coverage was discussed before identity was verified.
\begin{table}[t]
\centering
\begin{tabular}{lrrc}
\hline
\textbf{Edge} & \textbf{$e_i$} & \textbf{$s_j$} & \textbf{OK} \\
\hline
PID $\to$ CSV & 52 & 42 & $\times$ \\
\hline
\end{tabular}
\caption{Order violation example: Coverage can be 1 while OrderSafe is 0 if a child starts before its prerequisite finishes ($52\nless 42$).}
\label{tab:ordersafe-violation}
\end{table}

OIP--SCE keeps a single, auditable table with one row per phase. Two integers per phase the earliest start $s_j$ and earliest finish $e_j$ are sufficient to evaluate Coverage and OrderSafe. The configuration is validated up front, the evaluation is linear time, and the output is suitable for clinician-led review.

\section{Use Cases of OIP--SCE in Clinical Dialogues}
\label{sec:usecases}
OIP--SCE is designed to be authored by clinicians and operationalized by AI engineers. We illustrate this with two compact case studies and a pragmatic deployment workflow. Public, privacy-safe datasets of \emph{long-horizon, AI--Human clinical} dialogues are scarce. By contrast, the most-used clinical dialogue corpora are doctor--patient (human--human) or simulated for example, MEDIQA-Chat / MTS-Dialog and MDDial--and therefore are not logs of patients conversing with an AI system. \cite{abacha2023mediqa,mtsdialog_git,macherla2023mddial,liu2024survey} Large AI--Human resources do exist in the general domain (e.g., LMSYS-Chat-1M; ShareGPT), but they skew short on average and require filtering to find long, clinical threads. \cite{lmsys_chat1m,OpenGVLab_ShareGPT4o_2024}
To keep the evaluation recipe transparent and reproducible, we therefore demonstrate OIP--SCE on a public human--human clinical dataset (Case A) and on an anonymized AI--Human insurance call from our deployment setting (Case B).
\subsection{Case Study A: Respiratory History (doctor--patient)}
\label{sec:usecase_meditod}
We use \emph{MediTOD}~\cite{saley-etal-2024-meditod}, an English dataset of doctor-patient history-taking dialogues created from staged OSCE-style interviews (medical professionals role-play both sides) with comprehensive annotations; average dialogs are long (about 96 utterances) and cover respiratory and musculoskeletal complaints. We select MediTOD because real AI--Human clinical chat logs are rarely public for privacy reasons, whereas MediTOD is explicitly designed for research use and is available with sample JSON files.\footnote{\url{https://github.com/dair-iitd/MediTOD/blob/main/data/dialogs.json}}

\paragraph{Setup:}
An outpatient respiratory history follows a predictable pattern. The interview begins with the chief complaint (\texttt{SX\_DECL}), then establishes onset and duration (\texttt{SX\_ONSET\_DUR}), character of the symptom such as dry vs.\ productive cough and sputum color/volume (\texttt{SX\_CHARACTER}), and functional impact or progression (\texttt{SX\_SEV\_PROG}). Before broadening the scope, clinicians screen for urgent \emph{red flags} (e.g., hemoptysis, severe chest pain, high fever, presyncope; \texttt{RED\_FLAGS}) that can change urgency. They then take relevant background history (asthma/COPD or cardiac history; \texttt{PMH\_RELEV}), tobacco/vaping habits (\texttt{HABITS\_TOB}), and environmental exposures (pets/allergens, dust/fumes, occupational risks; \texttt{EXPOSURES}). Optional items often appear later: current respiratory medications and response (\texttt{MEDS\_ACTIVE}), family history (\texttt{FAMHX}), a plan for tests with rationale (\texttt{PLAN\_TEST}), and a provisional diagnosis (\texttt{DX\_PROV}). We annotate each of these as an OIP phase using the row schema in Sec.~\ref{sec:oip-sce}.

\paragraph{Order constraints:}
We enforce only the precedences that matter clinically. Chief complaint precedes the early detail phases (\texttt{SX\_ONSET\_DUR}, \texttt{SX\_CHARACTER}, \texttt{SX\_SEV\_PROG}); those precede \texttt{RED\_FLAGS}; red-flag screening precedes the background group (\texttt{PMH\_RELEV}, \texttt{HABITS\_TOB}, \texttt{EXPOSURES}); those precede the optional wrap-ups (\texttt{MEDS\_ACTIVE}, \texttt{FAMHX}, \texttt{PLAN\_TEST}, \texttt{DX\_PROV}). Only one edge is \emph{critical} for the order check here: \texttt{RED\_FLAGS} $\rightarrow$ \texttt{PLAN\_TEST} (do not propose tests before ruling out red flags). The remaining edges guide flow but are non-critical, so natural back-and-forth does not cause false failures.

\paragraph{Call-level decision:}
A dialogue passes if both conditions hold (Sec.~\ref{sec:oip-sce}): all required phases are completed (\textit{Coverage} = 1) \emph{and} every critical child starts after its parents finish (\textit{OrderSafe} = 1).

\paragraph{Mini slice (6 turns):}
\textbf{T0 Doctor:} What brings you in today? \\
\textbf{T1 Patient:} Cough getting worse for about two months, with some shortness of breath. \\
\textbf{T2 Doctor:} When did the cough start? \\
\textbf{T3 Patient:} Around two months ago. \\
\textbf{T4 Doctor:} Dry or productive-any sputum? \\
\textbf{T5 Patient:} Productive, white/yellow, maybe 5-10 teaspoons per day.

\paragraph{Row annotations (excerpt):}
\texttt{SX\_DECL}: $s{=}1,\ e{=}1,\ v{=}1$ ("cough + dyspnea");\quad
\texttt{SX\_ONSET\_DUR}: $s{=}2,\ e{=}3,\ v{=}1$ ("$\sim$2 months");\quad
\texttt{SX\_CHARACTER}: $s{=}4,\ e{=}5,\ v{=}1$ ("productive; white/yellow").\\
These rows (earliest start $s$, earliest valid finish $e$, verdict $v$, with evidence spans) are exactly what OIP--SCE consumes: \textit{Coverage} counts required phases with $v{=}1$, while \textit{OrderSafe} compares each critical parent's $e$ against its child's $s$. On this slice, only early phases appear, so Coverage is not yet complete but no critical edge can be violated.

\paragraph{Outcome on a full respiratory dialogue:}
On a complete MediTOD respiratory encounter using this phase map, all required phases finish in order; optional \texttt{MEDS\_ACTIVE}, \texttt{FAMHX}, \texttt{PLAN\_TEST}, and \texttt{DX\_PROV} also appear. Thus \textit{Coverage} $=1$ and \textit{OrderSafe} $=1$, so the call passes. This aligns with how MediTOD histories are conducted: the clinician drives a structured inquiry that maps cleanly to phases and keeps order checks meaningful.

\medskip
\noindent\emph{Notes on dataset choice:}
MediTOD's staged OSCE interviews (no real patient identifiers) provide realistic, multi-turn history-taking structure while remaining privacy-safe and publicly auditable, making it useful for demonstrating sequence sensitive checks like \textit{OrderSafe}. Although not AI--Human, OIP--SCE is model-agnostic and applies unchanged to AI--Human logs when available; we use MediTOD here to make the evaluation recipe transparent and reproducible.

\subsection{Case Study B: Insurance Benefit Verification (AI--Human)}
\label{sec:usecase_benefit}

\paragraph{Setting:}
As an early-stage healthcare entity, we applied OIP--SCE to a fully anonymized customer-service call handled end-to-end by an LLM-based patient-care specialist (PCS, "Agent") on our secure infrastructure. Audio was transcribed and de-identified prior to analysis: names, phone numbers, member IDs, dates of birth, and other HIPAA identifiers were removed or masked following 45~CFR~\S164.514(b). No individual-level information is reported here. \cite{ecfr164514}

\paragraph{Scenario:}
A patient-care specialist (PCS) agent calls an insurer's representative to verify benefits for a patient and two medications. The obligatory-information phases are: \emph{Patient Identification (PID)} - verify member identity (e.g., name, date of birth, ZIP, or member ID); \emph{Coverage Status Verification (CSV)} - confirm whether coverage is active or inactive; \emph{Drug Formulary Verification (DFV)} - for each medication, confirm formulary inclusion; \emph{Drug Restrictions Check (DRC)} - for each medication, record any prior authorization, step therapy, or quantity limits; \emph{Drug Copayment/Coinsurance (DCC)} - if no restrictions apply, obtain the copay or coinsurance amount; and \emph{Call Reference/Representative Name (CRN)} - capture the representative's name and the date or reference needed for audit.

\paragraph{Phase map and order:}
We use a simple DAG: \texttt{PID} $\rightarrow$ \texttt{CSV} $\rightarrow$ \texttt{DFV} $\rightarrow$ \texttt{DRC} $\rightarrow$ \texttt{DCC}; all precede \texttt{CRN}.
Critical edges for the order check are \{\texttt{PID}$\rightarrow$\texttt{CSV}, \texttt{DFV}$\rightarrow$\texttt{DRC}, \texttt{DRC}$\rightarrow$\texttt{DCC}\}.
Other edges are advisory (non-critical).

\paragraph{Requirement logic (branching):}
Let \(\mathrm{restrictions}=\) "exists a drug with PA/step-therapy/quantity-limit".
Requirements are:
\(\mathrm{req}_{\text{PID}}=\text{true}\);
\(\mathrm{req}_{\text{CSV}}=v_{\text{PID}}\);
\(\mathrm{req}_{\text{DFV}}=v_{\text{CSV}}\);
\(\mathrm{req}_{\text{DRC}}=v_{\text{DFV}}\);
\(\mathrm{req}_{\text{DCC}}=v_{\text{DRC}}\land\neg\mathrm{restrictions}\);
\(\mathrm{req}_{\text{CRN}}=\text{true}\).
We evaluate \textit{Coverage} using Eq.~\eqref{eq:coverage} and \textit{OrderSafe} using the critical-edge predicate from Sec.~\ref{sec:oip-sce}.

\paragraph{Short conversation slice:}

\textbf{T41-45 User:} "May I have the patient's name, date of birth, and zip code?" \\
\textbf{T46-52 Agent:} Provides name (spelled), DOB, and zip. \\
\textbf{T80-81 User:} "Coverage is active." \\
\textbf{T82-101} Formulary and restriction checks for two drugs (one PA required, one no restriction). \\
\textbf{T102-103} Copay quoted "\$25 after deductible." \\
\textbf{T107} Representative name and call date given as reference.

\paragraph{Row annotations (excerpt):}
\noindent\emph{Note:} $s_j$ is the earliest phase entry/attempt, and $e_j$ is the earliest valid completion. Thus a phase (e.g., CSV) can have $s_{\text{CSV}}<e_{\text{CSV}}$ when initiated earlier but confirmed later.\\
\texttt{PID}: \(s{=}45,\ e{=}52,\ v{=}1\) (identity fields confirmed); \quad
\texttt{CSV}: \(s{=}42,\ e{=}81,\ v{=}1\) ("coverage is active");\quad
\texttt{DFV}: \(s{=}82,\ e{=}99,\ v{=}1\) (both drugs on formulary);\quad
\texttt{DRC}: \(s{=}90,\ e{=}101,\ v{=}1\) (PA on drug~1; none on drug~2);\quad
\texttt{DCC}: \(s{=}102,\ e{=}103,\ v{=}1\) (copay \$25);\quad
\texttt{CRN}: \(s{=}104,\ e{=}107,\ v{=}1\) (rep name/date).

\paragraph{Outcome:}
All required phases passed, so \textit{Coverage} $=1$.
However, \texttt{PID}$\rightarrow$\texttt{CSV} fails the order test because \(\,s_{\text{CSV}}{=}42\) occurs before \(e_{\text{PID}}{=}52\) (premature coverage inquiry). The other critical edges are satisfied (\(\texttt{DFV}\rightarrow\texttt{DRC}\), \(\texttt{DRC}\rightarrow\texttt{DCC}\)).
Thus \textit{OrderSafe} $=0$ and the call does not pass overall:
\(\textbf{CallSuccess}{=}0\).
Clinically, this matches policy intent: identity must be completed before coverage discussion.

\paragraph{Notes:}
Because one medication had a restriction, \(\mathrm{restrictions}=\text{true}\) and \(\mathrm{req}_{\text{DCC}}=\text{false}\); the agent nevertheless obtained a copay, which is allowed but not required. In sites where copay is desired for \emph{any} unrestricted drug, the requirement rule can be tightened (e.g., require \texttt{DCC} if \emph{all} relevant drugs are unrestricted, or per-drug \texttt{DCC} rows); the OIP--SCE computation remains unchanged.

\section{Realizing Phase-Level Evaluation: Clinicians and Engineers, One System}
\subsection{Technical foundations: reliable, transparent, and scalable}
Phase-level evaluation only matters if it survives operations: ten thousand calls a day, shifting rules, staff turnover, tight budgets and the expectation that problems are not only detected but fixed. The division of labor is simple: clinicians define what good care requires; engineers make it observable, repeatable, and affordable.

Reliability at scale rests on \textbf{triage} and \textbf{evidence}. The validator clears routine, low-risk phases automatically and escalates only ambiguous or safety-critical cases. Every decision (pass or fail) points to a concrete span and a one-line rationale so reviewers can see what fired and correct it in seconds. This keeps misses and false alarms low where it matters (identity, consent, red flags) while aggregated views show which phases are drifting, which clinics struggle, and whether last week's fixes worked. When the assistant handles thousands of calls, this is how human attention stays on the small slice that truly needs judgment.

\textbf{Human-in-the-loop} is the control system, not a bolt-on. Escalations follow clinical risk (critical edges), uncertainty (low-confidence spans), and novelty (unseen phrasing). Clinician actions--confirm, override, nudge a boundary--feed a training queue and policy updates. If one phase chronically underperforms, we don't dig through transcripts; we strengthen that phase (better prompts, fine-tunes, new anchors, updated UI scripts) and watch its trend bend.

Staying current without burning engineering cycles requires \textbf{policy-as-code}. Clinicians own a versioned phase catalog (definitions, parents, critical-parents, acceptance examples) in a no-code editor. A linter catches cycles and gaps; changes run as canaries and shadow tests; one click re-audits last week's calls with the new rule. Regulation changes become catalog updates, not projects. The same structure keeps privacy and security first-class (de-ID where appropriate, encryption, access control, minimal data) and makes audits repeatable. Critically, phase granularity makes today's \textbf{turn-level metrics more useful}. Scoped to the right span, factuality checks live in \texttt{DX\_PROV}, empathy and clarity in \texttt{CONCERN\_DECL}, privacy filters in \texttt{PID}, and so on. Tools that are noisy over whole calls become precise inside the correct phase window false positives drop, and scores turn into phase-specific diagnostics clinicians can act on.

From a systems view, the pieces are straightforward: a clinician-owned \textbf{Phase Catalog}; an \textbf{Extraction Engine} (ASR, diarization, segmentation, anchored detectors); a \textbf{Validator} that implements Coverage and OrderSafe plus per-phase rubrics; an immutable \textbf{Evidence Store}; a \textbf{Clinician Console} for editing, review, and analytics; and MLOps/AIOps around it (model registry, drift monitors, cost/latency SLOs, canaries, error replay). Ops constraints are explicit: post-call results in seconds; real-time alerts only when clinically justified; compute bounded with high-precision anchors and small adjudicators; incident playbooks if anything regresses. Cost, latency, and reliability are treated as hard requirements, not afterthoughts.

\subsection{Clinical integration: governance, safety, and impact}
Governance is metrics, not vibes. We track quality (phase exact match, boundary error, Coverage and OrderSafe rates, per-phase precision/recall), efficiency (reviewer minutes per call, automation rate, cost per 100 calls), safety (critical-edge violations, PHI catches, override rates), and drift/fairness (phase failures by clinic, language, or age group). The table is bigger than ``AI + clinicians'': privacy/compliance, quality \& patient safety, call-center ops, legal/regulatory, health-IT/security, payers/providers, IRBs, patient advocates, and external auditors all sit with clinicians when phases are defined and when go-live criteria are set.

Our \textbf{compliance stance} stays simple and auditable. \textbf{Coverage} asks whether every required phase passed. \textbf{OrderSafe} asks whether each downstream phase began only after its upstream parents finished (strict on critical edges). We separate \textit{informational} from \textit{procedural} compliance (did the information appear vs. was it asked-then-answered), and we make volunteered-info policy explicit. Every verdict is evidence-backed; there is no silent credit.

A few hard problems remain and they are joint. Boundary detection in long, noisy calls still needs robust anchors and semi-automatic correction. Explanations must stay faithful and terse at volume. Plug-and-play phase schemas are needed across specialties, languages, and modalities (voice + text + attached documents). Real-time use should be introduced carefully (e.g., surfacing a missed red flag) to avoid alarm fatigue. Beyond model metrics, teams should track impact on audit cost, safety incidents, time-to-care, and patient experience.

\subsection*{Path Ahead:}
\begin{itemize}
    \item \textbf{Robust boundaries \& portable schemas:} reliable span detection in long/noisy calls, concise evidence-backed explanations, and phase catalogs that transfer across specialties, languages, and modalities.
    \item \textbf{Policy-as-code with safe updates:} versioned phase definitions, linted dependencies, canary + shadow promotion, one-click re-audit; compliance kept simple and auditable (Coverage, OrderSafe, and procedural vs.\ informational).
    \item \textbf{Systems \& ops SLOs:} seconds-scale post-call latency, bounded cost per 100 calls, drift monitors, error replay, incident playbooks; real-time alerts introduced sparingly and tied to critical edges to avoid alarm fatigue.
    \item \textbf{Shared ecosystem for progress:} multi-site datasets with phase spans and ask/answer flags, two public tasks (Boundary Detection, Compliance Verification), phase cards, and leaderboards that include adoption drivers--boundary exact match, critical-edge errors, escalation yield, cost, and latency.
    \item \textbf{Governance, privacy, and impact:} routine reviews with the full stakeholder set; de-identification, encryption, and least-privilege access by default; outcomes tracked beyond model metrics such as audit minutes, critical-edge violations, time-to-care, patient experience, and fairness slices.
\end{itemize}
Clinicians define the rules and examples; engineers make them fast, reliable, and affordable. With phase-specific evidence and trends in hand, the path from formalism to everyday practice becomes direct and testable.

\section{Conclusion}
Bringing AI into routine clinical practice demands more than high turn-level scores; it requires robust, transparent systems that clinicians and auditors can trust. Phase-level evaluation, as formalized in OIP--SCE, addresses persistent challenges in data quality, shifting regulations, and the messy reality of clinical workflows by structuring compliance as something clinicians can author and engineers can automate. Barriers remain: trust must be earned with clear evidence and usable tools, and systems must fit into real clinical routines, not just benchmarks. Yet by putting compliance logic and evidence in clinician hands, this approach helps AI move beyond technical promise toward everyday reliability. With OIP--SCE, we take a step closer to making AI in healthcare not only compliant, but truly usable and safe; bridging the gap between capability and clinical need, and paving the way for confident, clinician-driven adoption.
\bibliography{aaai2026}

\end{document}